\documentclass[a4paper,12pt]{article}
\usepackage{indentfirst} 
\usepackage{url}         
\usepackage{upquote}     
\usepackage{xcolor}      
\usepackage{orcidlink}   
\usepackage{fontawesome5}
\usepackage{tabularx}
\usepackage[framemethod=TikZ]{mdframed}
\usepackage[dvipsnames]{xcolor}
\usepackage{enumitem}
\usepackage{float}

\definecolor{githubgray}{HTML}{333333}
\newcommand{\githublink}[1]{
  \href{https://github.com/#1}{
    \textcolor{githubgray}{\faGithub}
    \hspace{0.5em}
    \texttt{#1}
  }
}

\usepackage{etoolbox}                       
\AtBeginDocument{\hypersetup{hidelinks}}    

\usepackage{authblk}                        

\usepackage{geometry}			
\usepackage{amsmath,amsfonts,amssymb,amsthm,mathtools}

\usepackage{graphicx}					
\graphicspath{{figures/}}		        
\usepackage{caption}                    
\usepackage{array,tabularx,tabulary,booktabs}	

\usepackage{cite}						

\usepackage{nomencl}                    
\begin{document}

\title{Exploratory Semantic Reliability Analysis of Wind Turbine Maintenance Logs using Large Language Models}
\author[1]{Max Malyi\orcidlink{0000-0002-1503-9798}\thanks{Corresponding author: \url{Max.Malyi@ed.ac.uk}}}
\author[1]{Jonathan Shek\orcidlink{0000-0001-5734-2907}}
\author[2]{André Biscaya\orcidlink{0000-0002-8158-4284}}

\affil[1]{Institute for Energy Systems, School of Engineering, The University of Edinburgh, Edinburgh, UK}
\affil[2]{Nadara, Lisbon, Portugal}

\date{August, 2025}

\maketitle

\begin{abstract}
A wealth of operational intelligence is locked within the unstructured free-text of wind turbine maintenance logs, a resource largely inaccessible to traditional quantitative reliability analysis. While machine learning has been applied to this data, existing approaches typically stop at classification, categorising text into predefined labels. This paper addresses the gap in leveraging modern large language models~(LLMs) for more complex reasoning tasks. We introduce an exploratory framework that uses LLMs to move beyond classification and perform deep semantic analysis. We apply this framework to a large industrial dataset to execute four analytical workflows: failure mode identification, causal chain inference, comparative site analysis, and data quality auditing. The results demonstrate that LLMs can function as powerful "reliability co-pilots," moving beyond labelling to synthesise textual information and generate actionable, expert-level hypotheses. This work contributes a novel and reproducible methodology for using LLMs as a reasoning tool, offering a new pathway to enhance operational intelligence in the wind energy sector by unlocking insights previously obscured in unstructured data.

\vspace{0.25cm}
\noindent\rule{\linewidth}{0.3pt}
\vspace{0.25cm}

\noindent\textbf{Keywords:} large language models, wind turbine reliability, maintenance logs, natural language processing, semantic analysis.

\vspace{0.25cm}
\noindent\rule{\linewidth}{0.3pt}
\vspace{0.25cm}

\noindent The project source code is hosted in an open source GitHub Repository:

\vspace{0.25cm}

\noindent \githublink{mvmalyi/llm-semantic-maintenance-logs-analysis}

\end{abstract}

\newpage

\tableofcontents

\makenomenclature
\nomenclature{LLM}{Large Language Model}
\nomenclature{O\&M}{Operation and Maintenance}
\nomenclature{SCADA}{Supervisory Control and Data Acquisition}
\nomenclature{WF}{Wind Farm}
\nomenclature{WT}{Wind Turbine}
\printnomenclature[2 cm]

\newpage

\section{Introduction} 
\label{sec:introduction}

Optimising operation and maintenance~(O\&M) strategies is essential for reducing the levelised cost of energy, as O\&M can account for a significant portion of a wind turbine's lifetime expenditure. To a large extend, such optimisation depends on robust reliability analysis, which is fundamentally reliant on the quality of operational measurements and logs of maintenance activities~\cite{carroll_failure_2016}. Although maintenance logs contain invaluable operational information, the wind industry lacks a standardised methodology for their analysis since the low-level details in such datasets are typically captured as unstructured free text with inconsistent terminology and formatting~\cite{hodkiewicz_cleaning_2016,hahn_recommended_2017}. Such logs from computerised maintenance management systems are primarily designed for operational tracking, not deep analytics; consequently, their descriptive text fields are not following a specific industrial standard, causing inconsistencies. The impact of this problem is substantial; foundational research has demonstrated that the choice of data cleaning techniques can alter key reliability parameters, like a component's characteristic life, by a factor of three~\cite{hodkiewicz_cleaning_2016}. This lack of inherent structure creates a significant barrier to automated analysis, thereby compelling operators to engage in time-consuming and costly manual processing to derive actionable insights~\cite{salo_work_2019}.

Historically, reliability studies have provided crucial findings by applying statistical models, such as those based on Poisson processes~\cite{alhmoud_review_2018}, or survival analysis to assess the impact of structured operational and environmental factors on turbine longevity~\cite{ozturk_assessing_2018}. While powerful, these quantitative methods often cannot leverage the rich, contextual detail embedded in historical data with technicians' descriptions of conducted maintenance. Even detailed statistical analyses of failure rates frequently rely on the manual interpretation of this maintenance log descriptions to correctly categorise events, a process that is both a bottleneck and a source of inconsistency~\cite{carroll_failure_2016}. Early attempts to automate text classification with traditional machine learning have shown promise~\cite{lutz_digitalization_2022}, but recent work has highlighted their limited ability to generalise across different sites. More fundamentally, it has been revealed that the manual gold standard labelling used to train these models can be highly subjective, with different expert groups producing failure rate key performance indicators that differ by over 300\% from the same dataset~\cite{walgern_impact_2024}.

The recent advent of large language models~(LLMs) represents a paradigm shift in natural language processing, offering unprecedented capabilities in understanding context and semantics. Unlike traditional models, modern LLMs can perform complex information extraction tasks in a zero-shot setting, though their application in industrial context can be constrained by the cost and data privacy concerns associated with proprietary models~\cite{walshe_automatic_2025}. The potential for these models in the wind energy sector is substantial, with recent studies identifying next-generation LLMs as the future frontier for standardising maintenance data and overcoming the limitations of current classifiers~\cite{walgern_impact_2024}. This trajectory extends beyond classification towards advisory roles, where LLMs could recommend repair actions based on operational data streams, provided that robust safety frameworks are in place~\cite{walker_safellm_2024}.

While our recent work has established a comprehensive framework for benchmarking the performance of leading LLMs on the foundational task of data classification and labelling~\cite{malyi_comparative_2025}, a gap remains in leveraging these models for more complex analytical tasks. This paper aims to address this by moving beyond classification to reasoning, presenting an exploratory framework that uses LLMs as an analytical tool for deriving  reliability insights directly from unstructured text. The contributions of this work can be divided into three parts: (1)~we introduce a flexible methodology based on structured prompt engineering to perform key reliability analyses, including failure mode identification, causal chain inference, and comparative site analysis, which can be adapted and extended for further tasks; (2)~we demonstrate that the insights generated by this semantic analysis provide a valuable qualitative complement to traditional quantitative reliability studies, and (3)~we provide an overview of the data quality audit that may be used as a reference for future standardisation of maintenance logs.

\section{Methods} 
\label{sec:methods}

This study employs an exploratory framework to assess the capabilities of LLMs for deriving reliability insights from unstructured natural language input used as descriptors in maintenance logs. The methodology encompasses two primary stages: (1)~the preparation of the dataset and the strategic selection of analytical cohorts, and (2)~the design and execution of specific analytical tasks using a structured prompt engineering approach.

\subsection{Data Preparation and Cohort Selection}

A preliminary pre-processed dataset consisting of 12152 maintenance logs went through an additional cleaning pipeline aimed at isolating the logs with potentially informative semantics in the descriptors. This resulted in a 10\% reduction, narrowing down the working dataset to 10,926 logs. These logs consist of work orders on corrective or preventive repairs and replacements from a portfolio of 25 onshore wind farms in Portugal, covering up to 7 years of operation (subject to a specific farm), and leveraging wind turbines of various types or manufacturers. The secondary data cleaning pipeline involved removing logs with uninformative descriptors or logs related to non-turbine infrastructure, cleaning the free-text fields using regular expressions to remove syntactic noise. To protect commercial sensitivity, all farm identifiers were anonymised using two-letter codes with a glossary generated for easy backwards references. In addition to such data as identifiers for logs, wind farms, turbines, subsystems, date, and age at event, the dataset has additional descriptive information that is not traditionally analysed. Two examples of such data that may be available in wind turbine maintenance logs are shown in Table~\ref{tab:data_sample}. 

\begin{table}[htbp]
\centering
\caption{Examples of descriptive information available in wind turbine maintenance logs.}
\label{tab:data_sample}
\footnotesize 
\begin{tabular}{p{4cm} p{5.7cm} p{5.7cm}}
\toprule
\textbf{Subsystem Name} & \textbf{Description} & \textbf{Observations} \\
\toprule
MV-Transformer & During electrical maintenance, damage was detected at the entrance of the medium-voltage busbar & None (\textit{can be empty in some logs}) \\
\midrule
Rotor Bearings & Lubrication of the Main Bearing & It is recommended to replace the lubricating grease for the main bearing \\
\bottomrule
\end{tabular}
\end{table}

Based on initial observations of statistical distributions within the dataset, specific data cohorts were selected to investigate distinct reliability questions. The rationale for each data subset is outlined further.

\begin{enumerate}
    \item \textbf{A Critical Subsystem:} The power converter module was chosen for failure mode analysis due to its relatively high frequency of maintenance events (1,065 logs), lack of hierarchical classification of specific issues occurring within the maintenance logs, and its direct relevance to the reliability of electrical systems being the focus of the project this study is a part of.
    \item \textbf{A High-Failure Turbine:} A single turbine exhibiting the highest normalised event frequency was selected for an in-depth causal chain analysis to test the LLM's ability to infer root causes from a chronological event sequence.
    \item \textbf{A Comparative Farm Group:} Three wind farms, further referred to as WF1, WF2, and WF3 were selected based on having a high and comparable number of maintenance logs. This cohort was designed to function as a natural experiment. \\ \textit{WF1 vs. WF2:} These farms are operating within the same site but use different turbine models – anonymised as WT Model 1 and WT Model 2, respectively. The basic technical characteristics are given in Table~\ref{tab:turbine_specs}. This comparison aims to isolate manufacturer-specific failure patterns. \vspace{0.25cm} \\ \textit{WF2 vs. WF3:} These farms use the same WT Model 2 turbine model but are situated in different geographical locations. This comparison aims to isolate site-specific factors driven by environmental or operational conditions.
\end{enumerate}

\begin{table}[htbp]
\centering
\caption{Technical characteristics of the wind turbine models used in the comparative analysis cohort.}
\label{tab:turbine_specs}
\begin{tabular}{lcc}
\toprule
\textbf{Characteristic} & \textbf{WT Model 1} & \textbf{WT Model 2} \\
\midrule
Wind Farms & WF1 & WF2, WF3 \\
Number of Turbines & 14 & 20 (WF2), 24 (WF3) \\
Rated Power (MW) & 2.5 & 2.5 \\
Rotor Diameter (m) & 100 & 90 \\
Hub Height (m) & 80 & 80 \\
\bottomrule
\end{tabular}
\end{table}

\subsection{LLM-based Analytical Framework}

\subsubsection{LLM Selection}
Two state-of-the-art LLMs were selected for this study: OpenAI's GPT-5 and Google's Gemini 2.5 Pro. These models were chosen for their advanced semantic reasoning and analytical capabilities, which are essential for the exploratory nature of the tasks at hand. GPT-5 was given the preference considering that it is a much more recent reasoning model excelling at benchmarks applied to this type of data \cite{malyi_comparative_2025}. Gemini 2.5 Pro has a significantly larger context window, which was beneficial for the data quality audit task requiring the analysis of the entire dataset.

\subsubsection{Task Design and Prompt Engineering}
A key component of this workflow is a structured prompt engineering. For each analytical task, the LLM was provided with a detailed prompt that assigned it a specific role (e.g., Reliability Engineer), provided rich context including the data and its description, outlined a clear list of tasks to follow, and specified a structured output format in JSON or Markdown to ensure the results were consistent and machine-parsable for further post-processing. This approach allows to apply the LLM as a specialised tool for each task. The four primary analytical workflows are summarised in Table~\ref{tab:prompt_framework}. All used prompts are shared within the open-source repository to make this exploratory study reproducible. 

\begin{table}[htbp]
\centering
\caption{Summary of the LLM-based analytical framework and prompt engineering design.}
\label{tab:prompt_framework}
\footnotesize
\begin{tabularx}{\textwidth}{@{} l p{4.5cm} X @{}}
\toprule
\textbf{Analytical Task} & \textbf{Objective} & \textbf{Key Prompt Engineering Elements} \\
\midrule
\textbf{Failure Mode Identification} & 
To synthesise unstructured logs for a single subsystem into distinct, quantified failure modes. & 
\textbf{Assigned role:} Reliability Engineer. \par
\textbf{Key tasks:} Grouping semantically similar events, providing descriptions, estimating log counts, and extracting supporting quotes. \par
\textbf{Required output:} A JSON object. \\
\addlinespace
\textbf{Causal Chain Inference} & 
To infer potential root causes by identifying causal links in a sequence of logs. & 
\textbf{Assigned role:} Diagnostic Engineer. \par
\textbf{Key tasks:} Reviewing the full sequence to identify plausible physical relationships, construct a hypothesis, and assess confidence. \par
\textbf{Required output:} A JSON object. \\
\addlinespace
\textbf{Comparative Site Analysis} & 
To identify and explain distinctive operational patterns by comparing logs from three different wind farms. & 
\textbf{Assigned role:} O\&M Analyst. \par
\textbf{Key tasks:} Identifying prevalent patterns at each site and formulating hypotheses based on the provided environmental and operational context. \par
\textbf{Required output:} A JSON object. \\
\addlinespace
\textbf{Data Quality Audit} & 
To assess the quality of entries and provide actionable recommendations. & 
\textbf{Assigned role:} Data Quality Expert. \par
\textbf{Key tasks:} Assessing text clarity, identifying common data issues, and providing actionable recommendations for technicians. \par
\textbf{Required output:} A Markdown report. \\
\bottomrule
\end{tabularx}
\end{table}

\subsubsection{Methodological Limitations}
It is important to acknowledge several limitations inherent to this exploratory study. First, the quality standard of the maintenance logs collected on wind farms currently remains relatively low within the industry. While the findings are illustrative of the framework's potential, they may not present universally generalisable failure statistics for the entire wind industry due to the quality of the data.

Second, the outputs generated by the LLMs should be interpreted as machine-generated inferences, not as established ground truth. They represent plausible starting points for investigation that require subsequent validation by domain experts with access to the full operational context of the assets.

Finally, while state-of-the-art LLMs possess powerful analytical capabilities, they are not infallible and are susceptible to risks such as hallucinations (generating plausible but incorrect information~\cite{walker_safellm_2024}). The structured prompt engineering methodology, which constrains the output format and requests supporting evidence from the source text, is designed to mitigate but not entirely eliminate this risk. The outputs should therefore be reviewed with critical expert oversight.

\section{Results} 
\label{sec:results}
The application of the LLM-based analytical framework yielded distinct insights across the four designed tasks. This section presents the outputs generated by the models with the full prompts and raw full-length results available in the accompanying public repository. The subsequent discussion section will interpret these findings in greater detail.

\subsection{Failure Mode Identification for Power Converters}
\begin{figure}[htbp]
    \centering
    \includegraphics[width=\textwidth]{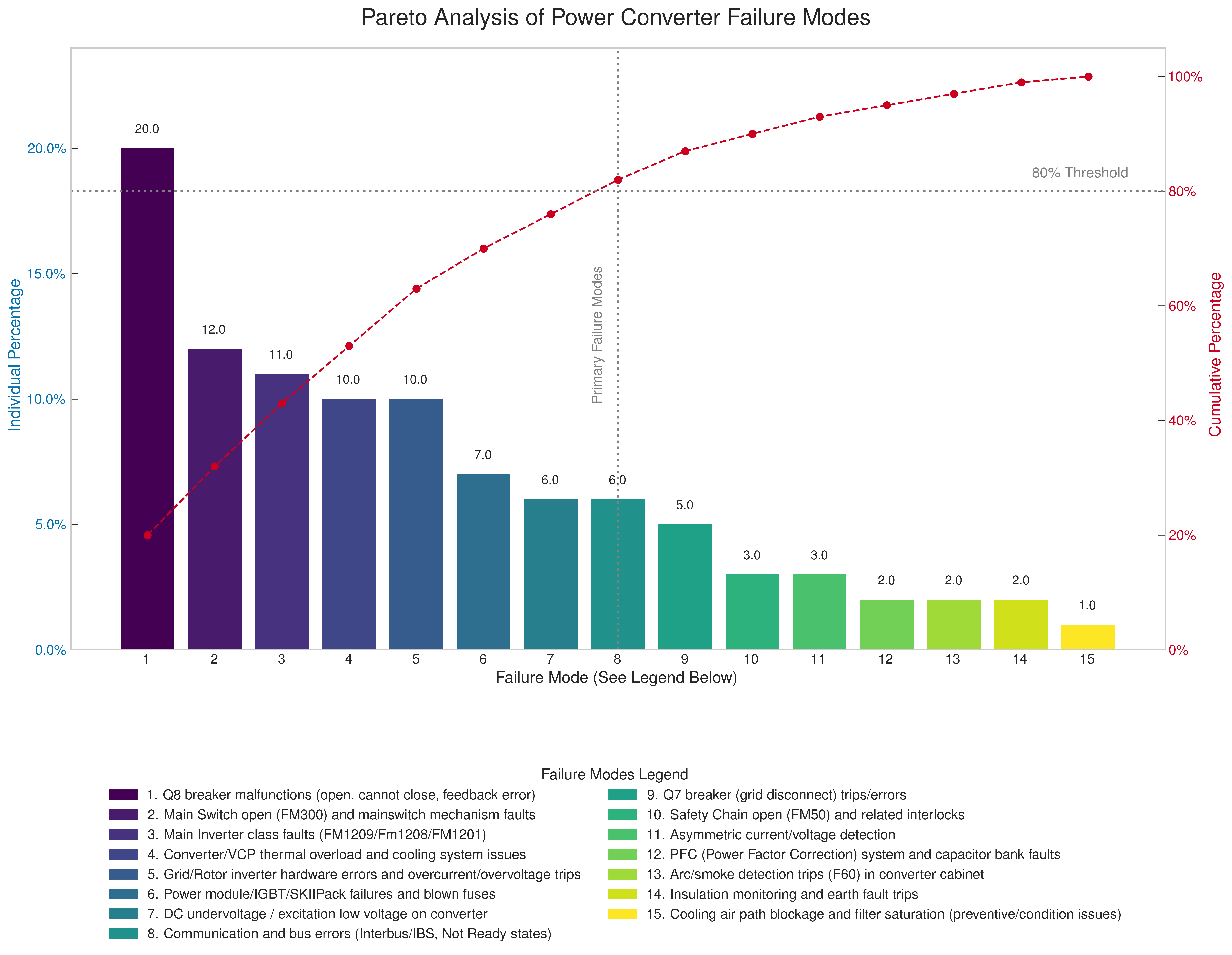}
    \caption{Pareto analysis of failure modes for the power converter identified and quantified by the LLM.}
    \label{fig:pareto_converter}
\end{figure}

The analysis of 1,065 maintenance logs specific to the power converter subsystem revealed 15 distinct failure modes. A Pareto analysis of these modes is presented in Figure~\ref{fig:pareto_converter}. The results show a highly concentrated failure distribution, consistent with the Pareto principle. The most frequent issue, Q8 breaker malfunction, accounted for around 20\% of all maintenance events in the observed subset. The top eight most common failure modes collectively represent around 80\% of all logged events, indicating that a small number of recurring problems may be the primary drivers of corrective maintenance for the power converter in this fleet.

The list of top eight synthesised failure modes and their technical descriptions is provided in Table~\ref{tab:failure_modes}, while a full list is available in the accompanying repository. The findings highlight the LLM's ability to not only identify high-level issues but also to categorise them with a high degree of technical specificity. The identified modes span a range of critical functions, including primary switchgear (e.g., Q8 breaker malfunctions), power electronic modules (e.g., Main Inverter class faults, IGBT failures), and auxiliary systems (e.g., thermal overload and cooling issues, communication bus errors). This level of granular detail, generated automatically from raw text, provides a comprehensive overview of the subsystem's reliability challenges.

\begin{table}[H]
\centering
\caption{Top 8 power converter failure modes synthesised by the LLM from maintenance logs.}
\label{tab:failure_modes}
\footnotesize
\renewcommand{\arraystretch}{1.5}
\begin{tabular}{r p{5.5cm} p{8.5cm}}
\toprule
\textbf{Rank} & \textbf{Failure Mode} & \textbf{Synthesised Description} \\
\midrule
1 & Q8 breaker malfunctions (open, cannot close, feedback error) & Frequent trips or command failures of the Q8 breaker due to feedback faults or mechanical/electrical issues cause the converter to disconnect or fail to reclose. \\
2 & Main Switch open (FM300) and mainswitch mechanism faults & The LV MainSwitch opens or fails to latch/close (often FM300) leading to converter de-energisation and repeated downtime, sometimes due to mechanical issues. \\
3 & Main Inverter class faults (FM1209/Fm1208/FM1201) & Main Inverter reports Class C/B/A faults indicating internal converter protection trips tied to hardware condition, insulation or thermal/operational thresholds. \\
4 & Converter/VCP thermal overload and cooling system issues & Over-temperature alarms on the VCP/PowerCab/Busbar and cooling system faults or clogged filters trigger converter derating or trips. \\
5 & Grid/Rotor inverter hardware errors and overcurrent/overvoltage trips & Hardware errors and protection trips on grid/rotor inverter stages (overcurrent, HW error, OVP) intermittently shut down the converter. \\
6 & Power module/IGBT/SKIIPack failures and blown fuses & Failures of power modules (SKIIPacks/phase modules) and IGBTs, often accompanied by blown fuses, require hardware replacement and cause trips. \\
7 & DC undervoltage / excitation low voltage on converter & Low DC or excitation voltage conditions on the converter DC link or rotor excitation path cause trips and not-ready states. \\
8 & Communication and bus errors (Interbus/IBS, Not Ready states) & Communication faults (e.g., Interbus FM700) and converter 'not ready' statuses interrupt inverter control and lead to stops. \\
\bottomrule
\end{tabular}
\end{table}

\subsection{Causal Chain Inference for a High-Failure Turbine}
\begin{figure}[htbp]
    \centering
    \includegraphics[width=\textwidth]{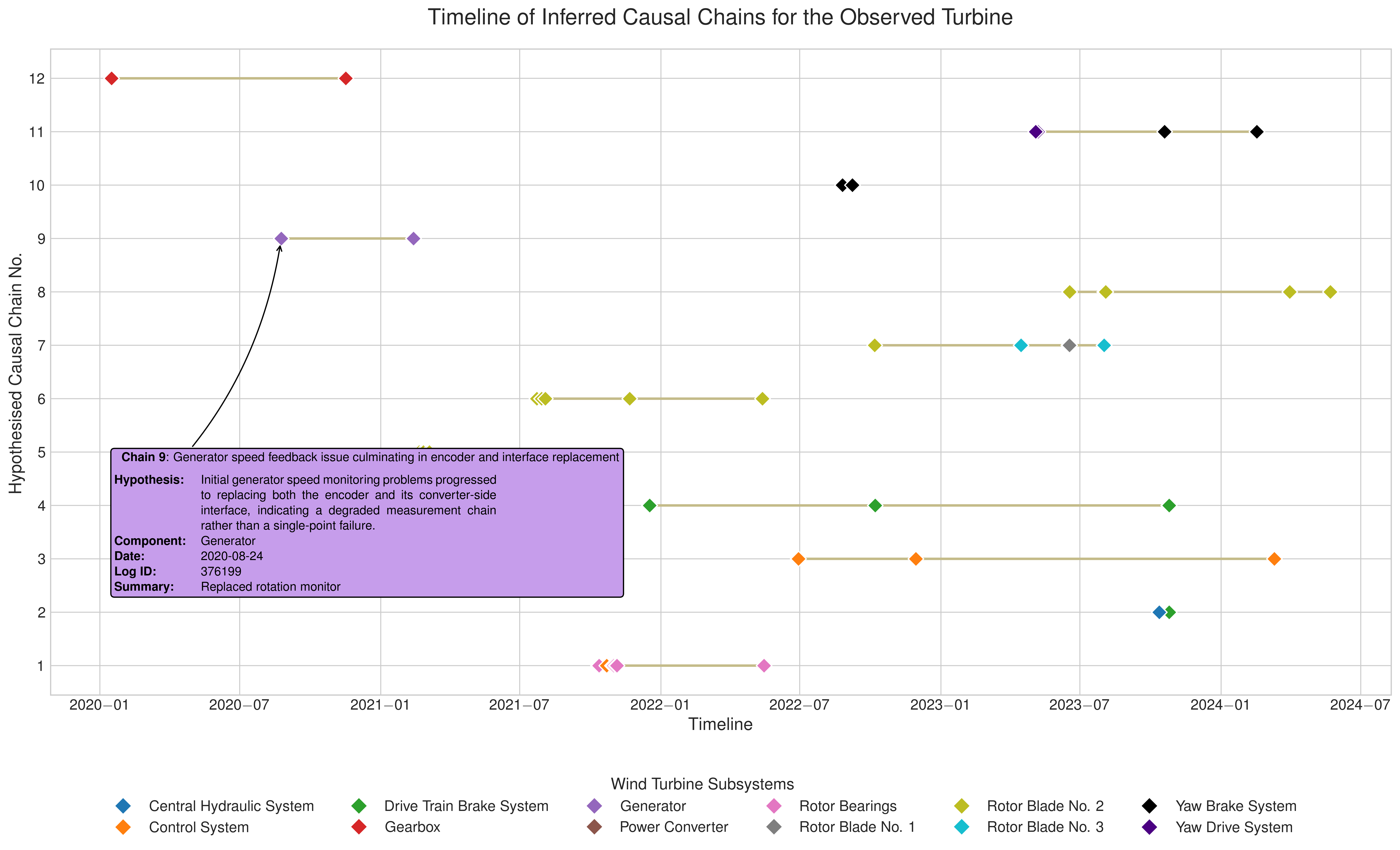}
    \caption{Timeline of the inferred causal chains for a single turbine, showing the potential temporal relationship between maintenance events.}
    \label{fig:timeline_turbine}
\end{figure}

To test the LLM's ability to infer root causes, a chronological sequence of 92 logs from a single turbine was analysed. The model identified twelve potential causal chains, each with a corresponding confidence level and inferred hypothesis. Figure~\ref{fig:timeline_turbine} visualises these chains over the timeline of turbine's available operational history. An interactive chart is available for viewing in the accompanying repository, providing an ability to see annotations for each inferred chain of events as in the example given, and zoom in to observe overlaying event markers.

The inferred chains varied in their explanatory power and homogeneity (consisting of the events from a single or multiple subsystems). One compelling, high-confidence example relevant to electrical systems was the on the generator speed feedback issue culminating in encoder leading to the interface replacement. The LLM identified a clear progression of related maintenance actions over several months: an initial replacement of a rotation monitor, followed by the replacement of the encoder interface board in the converter, and culminating in the replacement of the generator encoder itself. According to the model's hypothesis, such a sequence indicated a degraded measurement chain rather than a single-point failure, which is a plausible engineering diagnosis. In contrast, other assumed chains, such as one on yaw drive grease retention preceding brake adjustments were assigned a lower confidence level, suggesting a more tenuous correlation that would require significant expert validation. 

\subsection{Comparative Analysis of Manufacturer and Location Factors}
The comparative analysis of three wind farms identified distinct operational patterns, which the LLM linked to the contextual information provided regarding turbine manufacturer and location. The analysis successfully isolated both manufacturer- and location-specific patterns, which are detailed in Figure~\ref{fig:farm_comparison}.

\begin{figure}[htbp]
\centering
\begin{mdframed}[
    frametitle={LLM-Generated Comparative Analysis of Distinctive Failure Patterns},
    frametitlerule=true,
    frametitlerulewidth=0.4pt,
    frametitlebackgroundcolor=gray!20,
    backgroundcolor=gray!5,
    linewidth=0.4pt,
    roundcorner=5pt,
    innerleftmargin=10pt,
    innerrightmargin=10pt,
]
\footnotesize
\begin{itemize}[label={}, wide, labelwidth=!, labelindent=0pt, topsep=0pt, itemsep=0.2cm, leftmargin=*]
    \item \vspace{0.25cm} \textbf{WIND FARM 1} (WT Model 1; co-located with WF2)
    \begin{enumerate}[label=1.\arabic*., leftmargin=*, topsep=0pt, itemsep=0.3cm, after=\vspace{-\topsep}]
        \item \textbf{Pattern:} WT Model 1 converter ‘thread’-specific failures (repeated IGBT, fuse and MA contactor replacements across Threads 1–4). \par \hangindent=1.2em \hangafter=0 \textbf{Hypothesis:} Manufacturer-specific CCU architecture shows ageing-related thermal/electrical stress under frequent line events; component modularity leads to many thread-targeted repairs.
        
        \item \textbf{Pattern:} Axis-specific pitch system interventions (motors, brakes, batteries, chargers, and control modules). \par \hangindent=1.2em \hangafter=0 \textbf{Hypothesis:} Older WT Model 1 hub/pitch design with battery-centric actuation is sensitive to humidity ingress and ageing, driving recurrent replacements and tuning per axis.
        
        \item \textbf{Pattern:} Frequent anemometer icing and wind deviation state handling (de-icing and sensor swaps). \par \hangindent=1.2em \hangafter=0 \textbf{Hypothesis:} The local ridge/fog microclimate causes rime icing and measurement bias, necessitating repeated de-icing/maintenance.
    \end{enumerate}

    \centerline{\rule{0.5\textwidth}{0.2pt}\vspace*{1pt}\rule{0.5\textwidth}{0.2pt}}

    \item \textbf{WIND FARM 2} (WT Model 2; co-located with WF1)
    \begin{enumerate}[label=2.\arabic*., leftmargin=*, topsep=0pt, itemsep=0.3cm, after=\vspace{-\topsep}]
        \item \textbf{Pattern:} Widespread tower integrity and safety fixtures issues (corrosion, illegible signage, and emergency lighting failures). \par \hangindent=1.2em \hangafter=0 \textbf{Hypothesis:} Differences in tower coating specification and internal fixture quality for this WT Model 2 fleet, combined with site humidity, appear to drive accelerated degradation.
        
        \item \textbf{Pattern:} Systematic yaw drivetrain refurbishments (repeated gear replacements, brake relay swaps, and noise complaints). \par \hangindent=1.2em \hangafter=0 \textbf{Hypothesis:} The WT Model 2 yaw gearing and brake design may exhibit higher wear under local turbulence, and control tuning could increase actuation cycles.
        
        \item \textbf{Pattern:} High volume of fire detection sensor/base replacements across the site. \par \hangindent=1.2em \hangafter=0 \textbf{Hypothesis:} The installed detector model may be overly sensitive to dust or humidity in the tower environments, leading to false alarms and proactive replacements.
    \end{enumerate}

    \centerline{\rule{0.5\textwidth}{0.2pt}\vspace*{1pt}\rule{0.5\textwidth}{0.2pt}}
    
    \item \textbf{WIND FARM 3} (WT Model 2; different location)
    \begin{enumerate}[label=3.\arabic*., leftmargin=*, topsep=0pt, itemsep=0.3cm, after=\vspace{-\topsep}]
        \item \textbf{Pattern:} Recurring yaw brake/drive faults associated with misalignment errors and frequent caliper/pad work. \par \hangindent=1.2em \hangafter=0 \textbf{Hypothesis:} The complex terrain and directional shear at this location likely increase yaw misalignment events and braking duty, accelerating component wear.
        
        \item \textbf{Pattern:} MV cell component degradation (phase indicators, insulation boots, voltage signal lights). \par \hangindent=1.2em \hangafter=0 \textbf{Hypothesis:} Greater temperature swings and moisture ingress at this site may lead to dielectric and mechanical fatigue of MV cell components.
        
        \item \textbf{Pattern:} High prevalence of converter MI-class faults with frequent cooling system interventions. \par \hangindent=1.2em \hangafter=0 \textbf{Hypothesis:} Converter thermal management may be under greater stress from ambient conditions at this site, driving internal protection trips.
    \end{enumerate}
\end{itemize}
\end{mdframed}
\caption{Distinctive failure patterns and hypotheses synthesised by the LLM across the three wind farms. The framework successfully isolates both manufacturer- and location-specific factors by integrating operational and contextual data.}
\label{fig:farm_comparison}
\end{figure}

Furthermore, the analysis highlighted location-specific factors. Both WT Model 2 sites (WF2 and WF3) reported yaw system issues, but WF3, located in the more complex terrain of northern Portugal, showed a higher prevalence of yaw brake faults associated with misalignment errors, which the LLM hypothesised was due to increased braking duty under directional wind shear.

\subsection{Data Quality Audit and Model Comparison}
The final task involved using both GPT-5 and Gemini 2.5 Pro to audit the dataset for data quality issues and provide recommendations. Both models converged on the same core problems, but their outputs differed significantly in style and granularity, as detailed in the full audit reports in the accompanying open-source repository.

\subsubsection{Key Data Quality Issues Identified}
Despite their different analytical approaches, both models independently identified a consistent set of core data quality deficiencies. These issues, which are common in industrial maintenance records, significantly hinder automated analysis. The primary problems synthesised from the models' findings are:
\begin{enumerate}
    \item \textbf{Redundancy and Underutilisation of Fields:} A prevalent issue was the verbatim replication of text from the `Description` field into the `Observations` field. The `Observations` field was frequently underutilised, often containing redundant information, unhelpful placeholders, or being left empty, representing a missed opportunity to capture valuable contextual information.
    \item \textbf{Lack of Specificity and Quantification:} Many log descriptions were found to be overly general (e.g., "Falha de comunicação" -- Communication failure) and lacked the precision required for root cause analysis. The logs often referenced physical parameters like temperature or pressure without including specific numerical values or units.
    \item \textbf{Inconsistent Terminology and Formatting:} The audit highlighted a lack of standardised terminology across the dataset. Similar issues were described using different terms, and there were inconsistencies in language (mixing Portuguese and English), component naming, and the formatting of error codes, complicating data aggregation and automated parsing.
\end{enumerate}

\subsubsection{Actionable Recommendations for Standardisation}
Crucially, both LLMs generated a set of practical and actionable recommendations that, taken together, form a robust data governance strategy. The key suggestions for improving data entry standards included:
\begin{enumerate}
    \item \textbf{Implement Structured Data Entry:} Mandate a clear template that defines the purpose of each free-text field. For example, the `Description` field should concisely state the specific component and the observed problem, while the `Observations` field should be reserved for supplementary details like actions taken, parts used, and specific measurements.
    \item \textbf{Develop and Enforce Controlled Vocabularies:} Create a standardised glossary for common components of a lower hierarchy, failure types, and maintenance actions. Where possible, replacing free-text fields with drop-down menus based on this vocabulary would significantly improve data consistency.
    \item \textbf{Promote a Culture of Quantitative Reporting:} Train and encourage technicians to quantify their observations by including specific measurements and units (e.g., "temperature reached 85°C" instead of "overheating").
\end{enumerate}

\subsubsection{Divergence in LLM Analytical Styles}
An important methodological finding was the divergence in the analytical styles of the two models, despite being given the same prompt. GPT-5, even when analysing a 20\% subset due to its smaller context window, produced an exhaustive, highly granular report with ten specific issues and ten corresponding recommendations, suitable as a detailed checklist for a data governance team. In contrast, Gemini 2.5 Pro leveraged its larger context window to analyse the full dataset and generated a concise, high-level narrative, synthesising the problems into four overarching themes in the style of an executive summary for management. This finding suggests that the choice of LLM for such an audit can be a strategic one, depending on whether the desired output is a comprehensive, ground-level implementation plan or a high-level strategic overview.

\section{Discussion} 
\label{sec:discussion}
The results of this exploratory study demonstrate that modern LLMs can be leveraged as active analytical tools applied to natural language processing in different fields, including system reliability analytics. The framework demonstrated how such tools can perform complex semantic analysis offering a new approach to extracting intelligence from unstructured descriptors in maintenance logs.

\subsection{Interpretation of Analytical Findings}

\subsubsection{Synthesising Failure Modes and Causal Hypotheses}
The framework's ability to identify and quantify failure modes for the power converter highlights a primary strength of modern LLMs: synthesising coherent categories from textually diverse but semantically similar log entries. The model successfully aggregated over a thousand of disparate log descriptions to produce a classic engineering diagnostic tool, a Pareto chart (Figure~\ref{fig:pareto_converter}), that immediately focuses attention on the "vital few" issues, such as the recurring Q8 breaker malfunctions. However, it is important to address a key limitation. While the LLM excels at this semantic grouping, its direct numerical quantification of log counts and percentages in a zero-shot setting should be interpreted as a well-informed estimate rather than a precise calculation. This distinction underscores the ideal application of this framework: the LLM performs the complex, time-consuming task of semantic aggregation, while the human engineer validates the quantitative outputs, leading to the "co-pilot" concept.

Moving beyond static aggregation, the causal chain inference for the high-failure turbine demonstrates a more advanced reasoning capability. Here, the LLM analyses a chronological sequence of events to generate plausible, expert-level hypotheses about their potential interdependencies, complete with confidence levels to guide further investigation (Figure~\ref{fig:timeline_turbine}). The identification of a degrading generator speed measurement chain, for example, is a nuanced diagnosis that goes far beyond simple event correlation. The LLM does not provide a definitive root cause but rather a set of structured, testable hypotheses, which serves as an invaluable and accelerated starting point for a human reliability engineer's deep-dive analysis.

\subsubsection{Uncovering Site-Specific Reliability Signatures}
The comparative analysis of the three wind farms highlights the LLM's proficiency in semantic analysis, moving far beyond simple keyword matching. The model was able to discern subtle, technology-specific failure patterns, such as the converter `thread` issues unique to the WT Model 1 turbines at WF1, which contrasts with the yaw drivetrain issues identified at both WT Model 2 sites at WF2 and WF3. This demonstrates an ability to overcome the inconsistent terminology that plagues maintenance data and limits the generalisability of simpler classifiers across different sites. Furthermore, the LLM successfully linked these textual patterns to the provided environmental context. It hypothesised that anemometer icing at WF1 was due to the local ridge/fog microclimate common for the given location, and that the increased yaw brake faults at WF3 were related to the complex terrain and directional shear in the respective site. This capacity to integrate external context with semantic patterns in the source text is a powerful feature for uncovering site-specific reliability signatures that might be missed by purely statistical methods.

\subsubsection{Divergent Analytical Styles and Data Governance Insights}
The data quality audit revealed an interesting divergence in the analytical styles of the two state-of-the-art models. Despite converging on the same fundamental issues, data redundancy, lack of quantification, and inconsistent terminology, their outputs were tailored for different audiences. GPT-5 produced an exhaustive, highly granular report, presenting its findings as a detailed checklist suitable for a data governance team tasked with implementation. In contrast, Gemini 2.5 Pro provided a concise, high-level narrative, synthesising the problems into four overarching themes in the style of an executive summary for management. This suggests that the choice of LLM may depend not only on its technical capabilities but also on the desired depth and format of the analytical output. Critically, the audit confirms that the very data challenges highlighted by industry experts are prevalent in this dataset, reinforcing the need for advanced analytical frameworks like the one proposed, which are specifically designed to handle the inherent messiness of real-world industrial text data.

\subsection{LLMs as a Reliability Co-Pilot: Implications and Applications}

\subsubsection{A Paradigm Shift from Classification to Hypothesis Generation}
A key implication of this work is the potential for LLMs to function as analytical "co-pilots" for human experts, representing a paradigm shift in how textual O\&M data is utilised. Previous text mining efforts in the wind industry required either extensive manual work to create rigid, rule-based systems~\cite{salo_work_2019} or the use of traditional machine learning models to classify text into predefined categories~\cite{lutz_digitalization_2022}. While valuable, these approaches primarily aim to structure messy data. The workflows in this study, however, demonstrate a move beyond structuring to active synthesis and reasoning. In identifying failure modes, for instance, the model did not simply classify text; it grouped semantically similar descriptions, quantified their prevalence, and selected representative examples. Similarly, the causal chain inference shows the model's ability to form plausible hypotheses based on chronological and contextual clues. This capability to generate expert-level hypotheses and synthesise information represents a significant step towards the predictive and advisory roles envisioned for LLMs in critical industrial settings~\cite{walker_safellm_2024}. For this to be applied responsibly, a human-in-the-loop approach is essential, where the LLM accelerates and enhances the analytical process, but the domain expert remains the ultimate arbiter of the findings.

\subsubsection{Future Analytical Workflows and Opportunities}
This exploratory framework opens up several new opportunities for enhancing reliability analysis. The failure modes identified by the LLM, for example, can serve as a data-driven, semantic labelling hierarchy. Rather than relying on predefined component lists, operators could use this method to generate more granular failure categories (e.g., distinguishing between different types of converter breaker faults) directly from their own historical data. This enriched data could then be used to improve the accuracy of quantitative reliability models and overcome the subjective labelling issues that have limited previous classification efforts~\cite{walgern_impact_2024}. This can be applied at any scale, from a single high-failure turbine to an entire fleet.

Furthermore, the methodology's reliance on structured prompt engineering provides significant flexibility. Unlike a fine-tuned model trained for a single task, the LLM's role can be easily redefined to tackle other text-heavy analytical challenges. Operators could adapt this framework to analyse safety incident reports, compare the descriptive detail provided by different O\&M teams, or even triage incoming fault alarms based on the richness of their initial descriptions. Finally, the Data Quality Audit demonstrates a powerful application for data governance. The LLM-generated recommendations provide a concrete roadmap for improving data collection standards at the source. This creates the potential for a positive feedback loop: the framework is used to identify data quality gaps, which, when addressed, lead to higher-quality data collection. This improved data, in turn, enables more powerful and reliable LLM-driven insights, continuously enhancing the operational intelligence of the organisation.

\section{Conclusions and Future Work} 
\label{sec:conclusion}

This paper has introduced and demonstrated a novel, exploratory framework for leveraging the reasoning capabilities of modern LLMs to derive reliability insights from unstructured maintenance logs. Through structured analytical tasks, we have shown that LLMs can serve as powerful "reliability co-pilots" for human analysts. These models are capable of synthesising complex, semantically similar but textually diverse information to identify failure modes, inferring plausible causal chains from chronological data, and uncovering site-specific operational patterns by integrating external context. The framework provides a reproducible, human-in-the-loop methodology that moves beyond simple data classification to active hypothesis generation, offering a new pathway to unlock the immense operational value currently siloed in free-text maintenance records.

While this study establishes the potential of LLM-based semantic analysis, it also opens several promising avenues for future research.
\begin{enumerate}
    \item \textbf{Validation with Domain Expertise:} A critical next step is to validate the LLM-generated hypotheses, particularly the inferred causal chains and site-specific factors, through close collaboration with wind farm operators and reliability engineers. As the framework positions the LLM as a hypothesis generator, this expert validation, potentially augmented by correlating findings with other data sources like SCADA alarm sequences, is essential to confirm the accuracy of the insights.
    \item \textbf{Large-Scale Deployment and Generalisability:} The framework should be scaled and applied to larger, multi-operator datasets. This would not only test the generalisability of the prompt engineering strategies across different data collection cultures but also enable the construction of more comprehensive, industry-wide models of failure behaviour.
    \item \textbf{Integration into Predictive Maintenance Models:} The structured semantic insights generated by this framework, such as specific failure modes or precursor events from causal chains, are ideal candidates for features in quantitative predictive maintenance models. Future work should explore this fusion of semantic and time-series data, which could significantly enhance the accuracy of failure prediction algorithms.
    \item \textbf{Domain-Specific Model Fine-Tuning:} The availability of large, high-quality maintenance log datasets, improved through the data governance insights from this framework, would enable the fine-tuning of specialised models. This directly addresses the call for domain-specific LLMs for the wind energy sector  and would accelerate the development of expert, cost-effective analytical tools tailored to the nuances of wind turbine O\&M.
\end{enumerate}

\vspace{0.25cm}
\noindent\rule{\linewidth}{0.3pt}
\vspace{0.25cm}

\addcontentsline{toc}{section}{Acknowledgements}
\noindent\textbf{Acknowledgements}: This study is part of an ongoing PhD project on wind turbine reliability funded by the School of Engineering at the University of Edinburgh. Special acknowledgement is extended to Nadara, acting as the industrial partner, who provided the maintenance logs for this analysis. The authors also thank EDINA for providing access to OpenAI's models tested in the study.

\vspace{0.25cm}

\addcontentsline{toc}{section}{Data and Code Availability}
\noindent\textbf{Data and Code Availability}: To ensure the reproducibility and transparency of this study, the complete Jupyter Notebooks containing the data preparation pipeline, prompts, results, and analysis have been made publicly available in an open-sourced \textcolor{blue}{\href{https://github.com/mvmalyi/llm-semantic-maintenance-logs-analysis}{GitHub Repository}}. Please note that due to its sensitive nature, the maintenance log dataset used in this study cannot be shared. The open-source code is intended for operators and researchers to apply to their own proprietary data for semantic analysis of the descriptive columns in maintenance logs.

\phantomsection
\addcontentsline{toc}{section}{References}
\bibliographystyle{IEEEtran} 
\bibliography{maint-logs-semantic-analysis} 

\end{document}